\documentclass[11pt,a4paper]{article}
\usepackage{emnlp2020}
\usepackage{times}
\usepackage{latexsym}

\usepackage{verbatim}
\usepackage{amsmath}
\usepackage{multirow}
\usepackage{booktabs}
\usepackage{xcolor}
\usepackage{subcaption}

\usepackage{microtype}

\aclfinalcopy 
\def\aclpaperid{2982} 

\makeatletter
\renewcommand{\paragraph}{%
  \@startsection{paragraph}{4}%
  {\z@}{.2ex \@plus .2ex \@minus .2ex}{-1em}%
  {\normalfont\normalsize\bfseries}%
}
\makeatother

\usepackage{todonotes}

\newcommand{\Note}[2]{} 
\newcommand{\SideNote}[2]{} 
\renewcommand{\Note}[2]{\todo[color=#1,size=\small, inline=true]{#2}} 
\renewcommand{\SideNote}[2]{\todo[color=#1,size=\small]{#2}} %

\newcommand{\mg}[1]{\textcolor{black!35}{#1}}

\title{Content Planning for Neural Story Generation \\with Aristotelian Rescoring}

\author{Seraphina Goldfarb-Tarrant$^*$$^\dagger$,  Tuhin Chakrabarty$^*$$^\ddagger$,  Ralph Weischedel$^*$, Nanyun Peng$^*$$^\mathsection$ \\
  $^*$ Information Sciences Institute, University of Southern California \\
  $^\dagger$ University of Edinburgh, $^\ddagger$ Columbia University, $^\mathsection$ University of California, Los Angeles \\
  {\tt s.tarrant@ed.ac.uk}\textbf{,}
  {\tt  tuhin.chakr@cs.columbia.edu},\\
  {\tt weisched@isi.edu}\textbf{,}
  {\tt  violetpeng@cs.ucla.edu} \\}

\date{\today}

\begin{document}
\maketitle
\begin{abstract}
Long-form narrative text generated from large language models manages a fluent impersonation of human writing, but only at the local sentence level, and lacks structure or global cohesion. We posit that many of the problems of story generation can be addressed via high-quality content planning, and present a system that focuses on how to learn good \textit{plot structures} to guide story generation. We utilize a plot-generation language model along with an ensemble of rescoring models that each implement an aspect of good story-writing as detailed in Aristotle's \textit{Poetics}. We find that stories written with our more principled plot-structure are both more relevant to a given prompt and higher quality than baselines that do not content plan, or that plan in an unprincipled way.\footnote{Code at \url{https://github.com/PlusLabNLP/story-gen-BART}}
\end{abstract}

\begin{table*}[th!]
\centering
\small
\begin{tabular}{p{15.5cm}}
Prompt: Write the most beautiful end to the universe that you can imagine  \\
\hline
Naive Plot with Semantic Role Labelling Tags:   \\
{\tt \mg{<A1>} The universe \mg{<V>} end \mg{</s> </s> <A0>} ent 0 \mg{<V>} see \mg{<A1>} ent 3 \mg{</s> </s> <V>} dying \mg{<A1>} ent 1 \mg{\# <A0>} ent 1 \mg{<V>} left \mg{<A1>} ent 0 \mg{</s> <A0>} the last human \mg{<V>} live \mg{</s> <A1>} ent 6 \mg{<V>} end \mg{\# <A1}> ent 2 \mg{<V>} come \mg{</s> <A1>} the last one \mg{<V>} die \mg{</s> <A1>} a universe of life \mg{<V>} left } \\
\hline
Revised Plot with Semantic Role Labelling Tags:  \\
{\tt \mg{<A2>} The light \mg{<V>} filled \mg{<A2>} the sky \mg{</s>} \mg{</s>} \mg{</s>} \mg{<A2>} A bright flash \mg{<V>} lit \mg{</s>} \mg{</s>} \mg{\#} \mg{<V>} began \mg{<A2>} to grow \mg{\#} \mg{<A2>} the stars \mg{<V>} grow \mg{</s>} \mg{</s>} \mg{<V>} began \mg{<A2>} ent 0 \mg{<A2>} to fade \mg{\#} \mg{<A2>} The stars \mg{<V>} fade \mg{</s>} \mg{</s> <A0>} ent 2 \mg{<V>} looked \mg{<A2>} ent 1 me \mg{</s>} \mg{<V>} dying \mg{<A2>} star } \\
\hline
Naive Generated Story: \\
``You're dead. I'll be dead, and there'll always be no more of you."*  *``No, my dear. It's not true!" - The last of the world is dead, * * ``It'll never be over. \textit{\color{blue}The universe} will n't end. Not yet. But I'll see it, one way or another. There is no other way."*  *My last thought, the \textit{\color{blue}dying} man \textit{\color{blue}left} me. He told that I was \textit{\color{blue}the last human} to live. And he told me the universe would end, and the end will come soon. That your death would be \textit{\color{blue}the last one} to die, and the first one to be the most beautiful. * ``There is \textit{\color{blue}a universe of life} \textit{\color{blue}left}, my dear, and that's all there is.\\
\hline
Generated Story (from revision): \\
It was the end of the universe. \textit{\color{blue}The light} \textit{\color{blue}filled the sky}. The galaxy. All the light in all the worlds, and all the stars. \textit{\color{blue}A bright flash}. The first light. And then a bright flash of light... then a flash of darkness. Then a bright light. A flash of blackness and then the stars \textit{\color{blue}began to grow}. It was a bright, beautiful bright, light. And then, slowly the \textit{\color{blue}stars} started to \textit{\color{blue}grow} in number. There were millions and millions of stars.. But, slowly but surely, all the galaxies \textit{\color{blue}began to fade}. Slowly, but surely. Finally, it \textit{\color{blue}looked} at me with its eyes. One \textit{\color{blue}dying star}. \\
\hline
Gold Story: \\
It slept, and dreamt of twirling lights. It cradled countless twirling lights, all dancing in their flashing ways, dancing to the sound of a faraway flute. The light was vibrant and warm and spun all around It. The tiny lights grew with the tune, and more lights came to dance within the luminescence. It was surrounded by light, all waltzing in their ways to the distant music. It, too, swirled with the swelling symphony, watching the lights. There were lights of all different sizes and shades and brightness, but each was its own, unique light. Each one was brilliant. Each one was beautiful. All dancers to the Song... \\
\hline
\end{tabular}
\caption{Our proposed plot and story generation structure. We generate a Naive Plot, revise it with Aristotelian rescorers, then generate a story. In plots, $<$V$>$ denotes verbs while $<$A$\{0,1,2\}>$ denote arguments. ent $\{0...n\}$ are entities. We removed newline symbol $<$P$>$ from the generated stories and detokenized for better display.}\smallskip
\label{tab:first-examples}
\vspace{-1.5em}
\end{table*}

\begin{figure*}
  \centering
    \includegraphics[width=\textwidth]{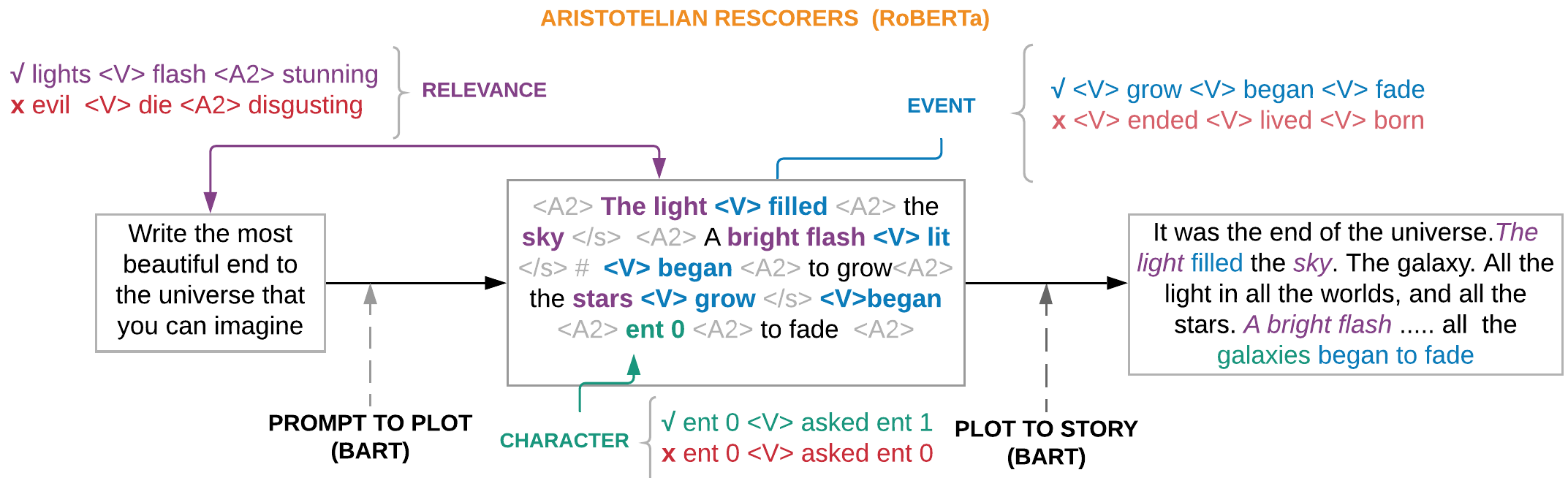}
    \caption{The full story generation pipeline. Base generation models in gray. Aristotelian rescorer input in color, with example positive \& negative training pairs. Plot structure shown with SRL tags for verbs and their arguments.} 
    \label{fig:sys_diagram}
    \vspace{-1em}
\end{figure*} 

\section{Introduction}
\label{sec:intro}
Despite many recent advances in Natural Language Generation, successful creative narrative composition remains elusive. 
Current neural approaches are plagued by difficulty in mastering structure, will veer between topics, and lack long-range cohesion. They successfully imitate the fluency and style of human writing, but on closer inspection sentences do not fit together to form a whole, and the reader is left with the impression that the generation has no content~\cite{see2019massively}.
This lack of structure also degrades the relevance of generations conditioned on a prompt or other source text - a strong language model will repeat key phrases from a given prompt but will not remain on topic. These issues are illustrated in the {\tt Naive Generated Story} in Table \ref{tab:first-examples}, where many of the sentences individually are fine, but do not fit together as one story, and do not all relate to the prompt. 

We hypothesise that this problem can be addressed with a focus on deeper latent narrative structures. In Aristotle's \textit{Poetics}, one of the most enduring treatises on the craft of writing good stories, the philosopher lays out the elements of story in order of importance. They are: 
\begin{enumerate}
    \setlength{\itemsep}{1pt}
    \setlength{\parskip}{0pt}
    \setlength{\parsep}{0pt}
    \item event choice and arrangement
    \item character
    \item relevant content\footnote{This directly translates as \textit{thought} in the original, but is explained to clearly mean the selection of relevant information}
    \item diction 
\end{enumerate}
An amateur masters skills later in the list, but mastery of event choice and event arrangement is what distinguishes a good writer \cite{aristotle1965loeb}. Next is character, then relevance, and only finally do style and diction matter. 

This philosophical framework fits remarkably well into the traditional Natural Language Generation Pipeline approach that emphasizes Content Planning \cite{reiter1997building}. 
The pipeline divides generation into three steps: \textit{Content Planning}, \textit{Microplanning} and \textit{Surface Realization}, where at each step input is modified and refined, getting closer to the final textual output. 
Incorporating a plot in order to generate stories can then be viewed as a proxy for  Content Planning/MicroPlanning before a language model makes use of it to convert it to a readable and grammatically correct natural language output (Surface Realization).

Inspired by both the Aristotelian and Content Planning Frameworks, we develop a novel system for story generation. We focus on developing a system that can learn to expertly select events, characters, and relevant content, and write good \textit{plot structures}. After the work on the plot is complete, a large language model can then do what it does best and fill in the descriptions, details, and local specifics of each story. 

For plot generation, we employ a few event-choice and event-arrangement rescoring models which assist in building the arc and cohesion of the plot, a character rescoring model that helps select which characters appear where, and a relevance model that is responsible for keeping the plot structure and the story on topic. As both improving plot-generation via rescoring and using an Aristotelian framework for neural generation are novel concepts, there is no previous work on how to implement them in practice.

Our contributions are: \textbf{1)} we  propose to leverage the principled Aristotelian framework for content planning, \textbf{2)} we propose an implementation of the framework using a revision-based approach via several rescoring models \textbf{3)} we show strong experimental results against 4 baselines. 

\section{Background}
\label{sec:bg}
Existing work in neural story generation has established the strength of adding a content planning stage to structure the generated content~\cite{yao2018plan,Fan2019StrategiesFS} (discussed in more detail in Section \ref{sec:related_work}). 
Specifically, this line of work trains a pipeline with one model that generates from \textit{prompt} $\rightarrow$ \textit{plot} and another that generates from \textit{prompt + plot}  $\rightarrow$ story.
It modifies the standard conditional generation task with a source $x = {x_1...x_n}$ (in this case, a prompt) and target $y = {y_1...y_n}$ (in this case, a story) to condition also on an abstract intermediate representation ${\bf z}$.
Note that the approach is not truly modelling $p(\textbf{y}|\textbf{x})$ since that would involve summing over all $\textbf{z}$. Instead, it models $p(\textbf{y,z}|\textbf{x)} = p(\textbf{z}|\textbf{x})p(\textbf{y}|\textbf{z},\textbf{x})$, but only shows the generated story $\textbf{y}$ at inference time. 

This is a more controllable task than open-domain generation conditioned on only a prompt $\textbf{x}$, provided that a good interim structure $\textbf{z}$ can be learnt. We follow this line and explore ways to improve  plot planning to close the gap between stories generated from gold plots and those from model-generated plots.

\paragraph{Plot Representation}
As there are no large datasets with parallel gold-standard plots and stories, all work on plot generation depends on silver-standard plots extracted from stories. We follow ~\newcite{Fan2019StrategiesFS} to represent plots in Semantic Role Labelling (SRL) format. We run coreference resolution to identify entities, and use a compression algorithm to discard less salient information.\footnote{Our representation is conceptually quite similar to that of \newcite{Fan2019StrategiesFS}, but is not identical, as their code is unavailable. We used AllenNLP \cite{gardner2018allennlp} to run the SRL model \cite{he2017deep} and Co-reference model \cite{lee2017end} and determined our own compression algorithm experimentally. Further details in Appendix \ref{sec:appendix_stopverbs}} 


\section{Approach} 

We focus on learning the best interim plot structure between the input prompt and the story surface realisation. 
As such, we learn the plot model $p(\textbf{z}|\textbf{x})$ and the story model $p(\textbf{y}|\textbf{z},\textbf{x})$ by fine-tuning a pretrained conditional language model BART~\cite{Lewis2019BARTDS} on 1) pairs of prompts and extracted silver standard plot-structures, and 2) pairs of prompts + plots and stories, respectively. Full implementation details can be found in Appendix \ref{subsec:appendix_impdet}. 

We propose to modify the decoding objective to incorporate input from each Aristotelian rescoring model $a$ $\in$ $A$ (the complete set of rescoring models, detailed further in Section \ref{subsec:rescore_model}) and re-rank the original, or ``naive" plot model hypotheses, bringing the plot representation closer to each rescoring model's specialty and desireable story attribute. A diagram of our final system in Figure \ref{fig:sys_diagram} shows each step of the generation process. The modified decoding objective becomes:

\vspace{-0.5cm}
{\small
\begin{align} \label{eq:decoding}
f_\lambda({\bf x, z}) = \sum_{i}^{m}-\log p(z|z<i,{\bf x}) + \sum_{j}^{|A|} \lambda_{j}a_j({\bf x}, z_{i...m}) 
\end{align}
\vspace{-0.5cm}
}\\
where $\lambda_j$ is the learned weight of the score given by $a_j$, as detailed in Section \ref{subsec:mix_weight}.



\subsection{Aristotelian Rescoring Models}
\label{subsec:rescore_model}
For all of our rescoring models, we train classifiers to \textit{distinguish} positive examples - the silver extracted plots - and negative examples, which are plots that are worse with regard to the aspect that we desire to encode, given the prompt ${\bf x}$. The intuition is that if the rescoring model can learn a particular principle, it can assist the plot-generation language model in creating content that encapsulates that principle. Mathematically, the classifiers learn $p(l|{\bf x, z}) = \frac{p({\bf x, z}, l)}{p({\bf x, z})}$, and we use the probability of the plot being a positive example (a more plausible plot) as our Aristotelian rescoring model:

\vspace{-0.6cm}
{ 
\begin{align} \label{eq:softmax}
a_j({\bf x}, z_{i...m}) = p_j(l_{pos} |{\bf x},  z_{i...m}).
\end{align}
\vspace{-0.6cm}
}\\
What differs for each model $a_j$ that specialises in a different Aristotelian aspect is the set of negative examples that we generated to capture the type of information it has learnt to discriminate, and the features it learns. We give more details about each Aristotelian rescorer as follows. Example simplified positive and negative pairs for each  are depicted in Figure \ref{fig:sys_diagram}. Full unsimplified pairs of positive and negative examples for each rescorer can be found in Appendix \ref{app:disc_training_examples}.

\paragraph{Event Rescorers.} 
The SRL extracted plots provide us with a structure that is very similar to event representations in event extraction literature. We thus consider an \textit{Event} to be composed of an action-based verb and its subject and object (a verb, subject, object tuple).\footnote{With modal verbs generally excluded in a stop list, which is included in Appendix \ref{sec:appendix_stopverbs}}
We experiment with three different ways to construct positive and negative event examples. SRL based plots are structured, and a random shuffle would be trivial to distinguish, so we need  more nuanced ways to learn good event choice and arrangement. We try:
\begin{itemize}
\vspace{-.3em}
\setlength\itemsep{-.3em}
    \item \textbf{inter-sentence shuffled events} we permute all sentences as a full chunk, and keep all events within a sentence together.
    \item \textbf{intra-sentence shuffled events} we permute the event tuples within a sentence, but keep each  verb and its arguments together.
    \item \textbf{verb-shuffled events} we permute only the event verbs within a sentence, leaving their arguments and contexts unchanged.
\end{itemize} 
The Event rescorer example in Figure \ref{fig:sys_diagram} depicts the verb-shuffled rescorer, where the model should learn that the progression \textit{grow $\rightarrow$ began $\rightarrow$ fade} is a suitable verb ordering, but that \textit{ended $\rightarrow$ lived $\rightarrow$ born} is unlikely. A simplified example for intra-shuffled from the same sample story would be the two phrases (within one sentence) {\tt \mg{<A1>} ent 12 \mg{<V>} shifted} $\rightarrow$ {\tt \mg{<A1>} ent 12 \mg{<V>} went \mg{<A2>} to sleep}, as opposed to the reverse ordering. For inter-shuffled it would be the two full sentences {\tt \mg{<A1>} ent 11 \mg{<V>} grew \mg{<A2>} louder} $\rightarrow$ {\tt \mg{<A1>} ent 11 \mg{<V>} fell \mg{<A2>} silent}, again as opposed to the reverse.

Each of these has specific strengths and weaknesses.
\textit{Inter-sentence shuffling} is closest to work on Narrative Event Chains \cite{Chambers2008UnsupervisedLO} and script-learning, which represent the fact that certain events are more likely to causally follow other events rather than precede them. However, since inter-sentence noising scopes globally over the entire plot structure, it is a harder task and may be difficult for the model to discriminate patterns.
\textit{Intra-sentence shuffling} is the same task but restricted to a more local scale, which makes the patterns clearer and more learnable but cannot capture long-distance Event Chains inter-sententially. It is also more sensitive to the style of a given story, as stories have a variable number of events per sentence.
Finally, \textit{verb-shuffling} focuses on verbs as the salient element of an event, and should teach both principles of verb ordering and of verb suitability for context, and avoid artifacts from reordering arguments.
However, since verbs are shuffled naively, the task can  in some cases be  too easy due to differences in verb selectional preferences.\footnote{For instance, an exclusively transitive verb could be swapped into the space of an intransitive one, and the edit would be trivially recognisable based on grammatical patterns.}

\paragraph{Character Rescorers.}
We represent character trajectory by distinguishing which character should appear at what point in the story.
We create training examples by taking each entity and all the preceding plot tokens up until the entity, and having the rescoring model choose between the true entity and a randomly sampled entity.\footnote{In other words, each training example is a section of plot sequence ${z_0...z_{t}}$ where $t$ is a randomly chosen entity, and the rescorer is given ${z_0...z_{t-1}}$ as context and must distinguish between the true and false $z_t$ entities that follow.) So the length of context varies for each training example.} The character rescorer must then distinguish between points in the plot when a pre-existing entity (a character already present in the story) should reappear, and if so which one, or whether a new entity should be introduced. The intuition is that this should encode typical patterns of a character's actions and relationships in particular contexts.

The Figure \ref{fig:sys_diagram} example encapsulates simplified correct character relationships (that a character should be asking something of another rather than of themself). This extends to much more complex character selection examples, as in the context: {\tt \mg{<A0>} ent 0 \mg{<V>} saw \mg{<A1>} the light of a campfire \mg{</s> <A1>} ent 2 \mg{<V>} laying \mg{<A2>} there \mg{</s>}
\mg{\# <A1>} light \mg{<V>} bouncing \mg{</s> <A0>} ent \underline{?}} where the positive is {\tt 0} and the negative is {\tt 3}. Here the difference between the positive and negative captures whether or not it is natural to introduce a new character at this juncture (it is not, as in the gold plot a new character does not appear till a few sentences later).

\paragraph{Relevance Rescorers.}
We take an approach inspired by prior work on learning to discriminate between random and true continuations of story sentences \cite{holtzman2018learning}. We consider pairs of prompts and plots, where a positive example is the true plot and the negative is a randomly selected plot from elsewhere in the training data. This prompt and plot pairing is a much more difficult task than pairing context and continuation sentences, since they are less closely connected. Once trained, this rescorer is expected to tell which kinds of plot words, verbs, and SRL patterns belong with which kinds of prompts. The example in Figure \ref{fig:sys_diagram} shows a relevant and irrelevant sentence for the given prompt. During training, the rescorer will receive the entire gold story and an entire random story to match to a prompt (see Appendix \ref{app:disc_training_examples}).


\subsection{Rescoring Model Architecture}
\label{subsec:disc_arch}
There is an inherent tension in training a useful rescoring model: discrimination tasks for which even simple models can perform well may have inherent data artifacts and therefore not be helpful for modeling plots. However, discrimination tasks that are so hard that all models have low accuracy are also likely to be unhelpful.
We experiment with three different architectures for our rescorers. We start with ngram-baseline models\footnote{Implemented in Scikit-learn \cite{scikit-learn}} to better judge the baseline difficulty of a given task and take artifacts of data creation into account. This is more informative than random chance accuracy. We also experiment with augmented versions of the CNN architectures used in \newcite{holtzman2018learning}, and RoBERTa models \cite{liu2019roberta}, and find RoBERTa to have the best performance for each Aristotelian concept.
\begin{enumerate}
\vspace{-.3em}
\setlength\itemsep{-.3em}
    \item \textbf{XGBoost with ngrams:} We used n-grams in range of (1,4) as features and trained a XGBoost model with 200 gradient boosted trees with a maximum tree depth of 5. 
    \item \textbf{CNN with maxpooling:} We used a CNN-based architecture \cite{holtzman2018learning} but augmented it with BART position and subword encodings because our event tasks are ordered, so pooled or averaged representations that don't represent permutations differently would be indistinguishable.   
    \item \textbf{RoBERTa-large} \cite{liu2019roberta} has shown excellent performance on various sentence pair classification tasks. We expect this large pre-trained language model to be more efficient in being able to discriminate between a well-formed sequence of words and a poorer one. To this end we finetune RoBERTa-large with a classification specific final layer as the final option to build rescorer models.
\end{enumerate}
Accuracies for different rescorer architectures by aspect are shown in Table \ref{tab:disc_acc}. As we hypothesised from the nature of many of the tasks (Section \ref{subsec:rescore_model}), the inter-sentence shuffled task is more difficult because the noising is global. This is reflected in the barely above chance scores of the ngram-baseline. The Verb-shuffling high ngram-baseline performance shows that our suspicions about this task being easier were also correct. Intra-shuffling was the only surprise, and turned out to be more difficult than we expected and to have the largest gap between baseline ngram and CNN performance. RoBERTa scores are high across the board, so we use RoBERTa for all models in the final system.

\begin{table}[t]
\centering
\small
\begin{tabular}{|l|l|l|l|l|}
\hline
\multicolumn{2}{|l|}{\quad Plot Elements} & XGB & CNN   & RoBERTa       \\ \hline
\multicolumn{2}{|l|}{Relevance}  & 55.2 & 56.0 & \textbf{89.3} \\ \hline
\multicolumn{2}{|l|}{Entity}       & 64.1 & 76.0 & \textbf{92.3} \\ \hline 
\multirow{3}{*}{Event} & Inter-Shuffled   & 55.4 & 53.0 & \textbf{96.6} \\ \cline{2-5}
& Intra-Shuffled  & 58.5 & 75.0 & \textbf{97.3} \\ \cline{2-5}
& Verb-Shuffled & 76.1 & 75.0 & \textbf{95.8} \\ \hline
\end{tabular}
\caption{Accuracy for training Rescorers using three different architectures.}
\label{tab:disc_acc}
\vspace{-1.5em}
\end{table}

\subsection{Mixture Weight Training and Ablations}
\label{subsec:mix_weight}
We learn optimal weights for rescorers online during decoding using a held-out validation set $V$, and use these weights during inference via sampling. We minimize Margin Ranking Loss of the negative log probability of each validation sample between the gold (${\bf z}$) and hypothesised (${\bf \hat z}$) plot structures.

\vspace{-0.5cm}
{\small
\begin{align} \label{eq:mix_training}
\underset{\lambda}{\mathcal{L}_{mix}}= \sum_{{\bf x,z} \in V} \sum_{i=1}^n \max(0, -(f_\lambda({\bf x,z}_{<i}) - f_\lambda({\bf x},\hat{\bf z}_{<i})) 
\end{align}
\vspace{-0.5cm}
}\\
where $i$ indexes the word position, $n$ denotes the plot length, $f_\lambda$ is the same as in Equation \ref{eq:decoding}, and we are training the $\lambda$ weights with this objective.\footnote{We experiment with learning on each token (in which case, each validation sample will contain a few hundred updates) vs. on each complete sample (an entire plot). We find both approaches to perform similarly (with every-token gaining $1\%$ accuracy) and use an every-token approach in for our final system. Mixture weight hyperparameters can be found in Appendix \ref{sec:appendix_mix_weight}} 
We train mixture weights both for combinations of rescorers and for ablations using each rescorer individually, to isolate the contribution of each one.

Mixture weight training accuracy is in Table \ref{tab:ablation},\footnote{Our 4-scorer ensemble subtracts out the Intra-event model as it is conceptually a more local version of the Inter-event model, and has a lesser effect on Ranking Accuracy} which we report as Ranking Accuracy, the number of samples where the generation has higher probability than the gold. There we also include our automated plot metrics on the validation set for each ablation (further detail on those metrics in Section \ref{subsec:metrics}). As Table \ref{tab:ablation} shows, Inter-event is the strongest of the individual rescorers, though all five together achieve the best performance. This seems to indicate that each method of creating negative event examples is encoding a separate helpful piece of information, rather than one of them alone being the best approach.

Given that the ensemble of many rescorers was best, this suggests further investigation into the interactions between rescorers, and into whether there is a pareto front and in which cases certain rescorers dominate. Due to the difficulty of human evaluation of long stories and the accompanying funding constraints, we leave this as a suggestion for interesting future work to analyse. We thus select the ensemble of all 5 rescorers with optimal learned mixture weights for our final Aristotelian Plot System.

\begin{table}[t]
    \centering
    \small
    \begin{tabular}{l|l|l|l}
         Rescorer & RA $\uparrow$ & V:T $\uparrow$ & E $\uparrow$  \\
         \hline
         All 5 & \textbf{0.43} & \textbf{2.07} & 7.15 \\
         \hline
         All 4 (-intra) & 0.35 & 1.72 & 8.78 \\
         \hline
         Inter-event & 0.37 & 1.76 & 8.87 \\
         \hline
         Intra-event & 0.16 & 2.01 & 6.58 \\
         \hline
         Verb-event & 0.15 & 1.66 & 8.89 \\
         \hline
         Entity & 0.17 & 1.68 & \textbf{9.18} \\
         \hline
         Relevance & 0.17 & 1.71 & 8.89 \\
         \hline
    \end{tabular}
    \caption{Ranking accuracy (RA) and generated plot metrics i.e Verb:Token ratio (V:T), Entities(E) for rescorer ablations}
    \label{tab:ablation}
    \vspace{-1.5em}
\end{table}


\section{Experimental Setup}
\paragraph{Dataset.}
We use the Writing Prompts dataset \cite{fan2018hierarchical}, which is a large collection of user-generated 
stories along with their associated prompts from Reddit, to benchmark our models. It is particularly suited to this task since it is both hierarchical (contains pairs of titles and stories, which enables the use of a plot as an interim step) and contains many diverse long-form stories that are very challenging to learn to structure.\footnote{Dataset statistics are in Appendix \ref{subsec:dataset_stats}}

\subsection{Baselines}
We compare our generations to the two strongest recent story generation systems as well as two ablated versions of our own system.
\paragraph{Targeted Common Sense Grounding Model.} \citet{mao2019improving} propose a multi-task learning scheme to achieve quantitatively better common sense reasoning in pre-trained language models by leveraging auxiliary training signals from datasets designed to provide common sense grounding. 


\paragraph{Knowledge-Enhanced Commonsense Model.} \citet{guan2020knowledge} devise a knowledge-enhanced pretraining model for commonsense story generation leveraging external knowledge bases on the ROCStories dataset \cite{mostafazadeh2016corpus}.  To further capture causal and temporal dependencies between sentences in a reasonable story, they employ multi-task learning which combines a discriminative objective to distinguish true and fake stories during fine-tuning. 


\paragraph{Prompt to Story.} This fine tunes the BART model directly with the prompt and story pairs without access to a plot structure.

\paragraph{Naive Plot.} This utilizes a plot structure to write a story, but does not incorporate the Aristotelian Rescorer ensemble.

We do not compare to the strong system of \newcite{Fan2019StrategiesFS}, as they do not release their code and the authors were unable to make their output available after being contacted.\footnote{Code for \newcite{fan2018hierarchical}, which does \textit{not} use plots, is available but is trivially beaten by any recent story generation system.} However \citet{guan2020knowledge} is one of our baselines and achieves better results on \newcite{mostafazadeh2016corpus} than their approach. Our \textbf{Naive Plot} baseline is most comparable and quite similar to their system, save that it lacks their verb attention, but leverages powerful BART pretraining. 



\subsection{Metrics}
\label{subsec:metrics}
We use a combination of human and automatic metrics to evaluate all systems. Automatic metrics are evaluated on 1000 randomly selected prompts and their associated plot structures and stories from the WritingPrompts test set. As automatic metrics are known to be overly coarse for open domain NLG \cite{novikova2017} we also report human evaluation results on 110 stories, with each story evaluated by 3 separate judges.\footnote{Workers were paid \$12/hr. They were given content instructions and told to disregard punctuation and spelling. Further details on compensation and instructions in Appendix \ref{sec:appendix_mturk}.} All metrics are averaged over all stories within the evaluation set.
\paragraph{Plot Structure Metrics.} For plot structures we use purely automated metrics, as it is difficult for humans to read the abstract interim representation of SRL tags and entity anonyimisation:
\begin{itemize}
    \itemsep-.4em 
    \item Vocab:Token Ratio
    \item Entities per plot
    \itemsep-.2em 
\end{itemize}
We choose to focus on these metrics as indicative of performance as they show the largest gap between gold and generated stories. \textit{Vocab:Token Ratio} is commonly used as a measure of originality and diversity of content for stories, and can be readily applied to plots. \textit{Entities per plot} we found to be a good metric as a reasonableness check, and for ablations, as weak models tend to have too few entities. Both metrics also relate to repetition, a known weakness of neural generated stories in comparison to human stories even with large pretrained transformers.  

\paragraph{Automatic Story Metrics.}
\begin{itemize}
    \itemsep-.4em 
    \item Vocab:Token Ratio
    \item Unique Verbs \& \% Diverse Verbs: \% of verbs not in the top 5 most frequent\footnote{We report these for comparability to \newcite{Fan2019StrategiesFS}; as they do we identify verbs via \url{https://spacy.io/}}.
    \item Intra-story Trigram Repetition: trigram repetition \textit{within} a story
    \item Inter-story Trigram Repetition: trigram repetition \textit{between} stories
    \itemsep-.2em 
\end{itemize}
We again use Vocab:Token ratio as a rough diversity metric. We report also inter-story trigram repetition rates and intra-story trigram repetition rates \cite{yao2018plan}. The former is a diversity metric - if stories look fine but inter-story repetition is high, it means the language model has learned to tell only very similar stories even when conditioned on diverse prompts. Intra-story trigram repetition is a fluency metric, and measures the proportion of trigrams within a single story that are repeated. 

For all of these metrics, there is a tension between diversity metrics, which bring a generation closer to human quality, and fluency metrics, which can degrade as diversity increases.

\paragraph{Human Metrics.}
We run two separate experiments to measure improvement in our target areas of \textsf{Relevance} and \textsf{Overall Quality}. \textsf{Relevance} is defined as whether stories both relate to the given prompt and remain on topic for the duration of the story. \textsf{Overall Quality} is defined to be combination of \textit{coherence}, \textit{interestingness}, and \textit{relevance}, similarly to most other story generation work.\footnote{Surveys used for both may be found in Appendix \ref{sec:appendix_mturk}, as well in in the code repository.}
We postprocess stories for human review by detokenizing\footnote{We use MosesDetokenizer \cite{koehn2007moses}}, removing special end-of-sentence tokens, and truncating to 250  (whitespace separated) words. 
We have Mechanical Turk workers rate all systems' outputs on the same prompt comparatively on a \textit{likert score} (1-5) across both metrics. But since likert scores are well known to exhibit a central tendency bias, it is likely to be unreliable on distinguishing between systems that are close in performance, particularly as reading 5 long generations introduces significant cognitive load.
Therefore, we further conduct \textit{pairwise comparisons} between the top 3 systems in the likert experiment. 


\paragraph{Test Data Selection.} In contrast to previous work on this dataset, our 110 human titles are randomly sampled from a \textit{filtered version} of the test set. Writing Prompts has a one-to-many relationship between the prompts and stories.\footnote{Of the 303358 prompts, only $1/3$ (107665) are unique} The dataset also contains an artifact of the sort of topic that is upvoted on reddit\footnote{Mostly aliens}, so many test prompts are minor variations on the same topic. We hypothesise that some of the gap between reported performance in papers on this dataset and performance in the wild is due to the artificially high similarity between training and test prompts, so we randomly sample from the test set, but exclude prompts with extremely high lexical overlap with training prompts.\footnote{72\% of prompts were excluded. We used sequencematcher \url{https://docs.python.org/3/library/difflib.html} and spacy vector similarity \url{https://spacy.io/} to exclude prompts with a similarity of 1 to any prompt in the training data when stopwords are removed. The two systems gave identical results.} 


\section{Results}
\label{sec:results}

\begin{table}
\centering
\small
\begin{tabular}{l|l|l|l@{\ }}
System & Voc:Tok ratio & Entities & Avg Tok  \\
\hline
Naive Plot & 1.52 & 8.25 & 199 \\
\hline
Aristotelian Plot & 1.81 & 7.49 & 168 \\
\hline
Gold Plot & 3.59 & 9.26 & 371 \\
\hline

\end{tabular}
\caption{Metrics for plots  }\smallskip
\label{tab:plot-results}
\vspace{-1em}
\end{table}

\begin{table}
\centering
\small
\begin{tabular}{@{\ \ }l@{\ \ }@{\ \ }p{0.7cm}@{\ \ }@{\ \ }p{1.1cm}@{\ \ }@{\ \ }p{0.8cm}@{\ \ }@{\ \ }p{0.8cm}@{\ \ }}
\hline
 & \multicolumn{4}{c}{\textbf{Automatic Evaluation}} \\ 
System & Voc: Tok$\uparrow$ & Diverse Verb \%$\uparrow$ & Intra-Rep$\downarrow$ & Inter-Rep$\downarrow$ \\
\hline
Mao et al. (2019) & 6.6 & 81.8 & 5.68 & 27.5  \\
\hline
Guan et al. (2020) & 2.3  & 71.9 & 0.60 & 56.1  \\
\hline
Prompt-to-Story & 1.5  & 68.9 & 0.22 & 65.1  \\
\hline
Naive Plot & 1.4  & 76.4 & 0.11 & 63.8    \\
\hline
Aristotelian Plot & 1.5  & 74.8 & 0.12 & 64.1  \\
\hline

\end{tabular}
\caption{Automated metrics for all models.}\smallskip
\label{tab:story-results}
\vspace{-1.5em}
\end{table}

\begin{figure*}[t]
  \centering
  \begin{subfigure}[t]{0.5\textwidth}
    \includegraphics[width=\textwidth]{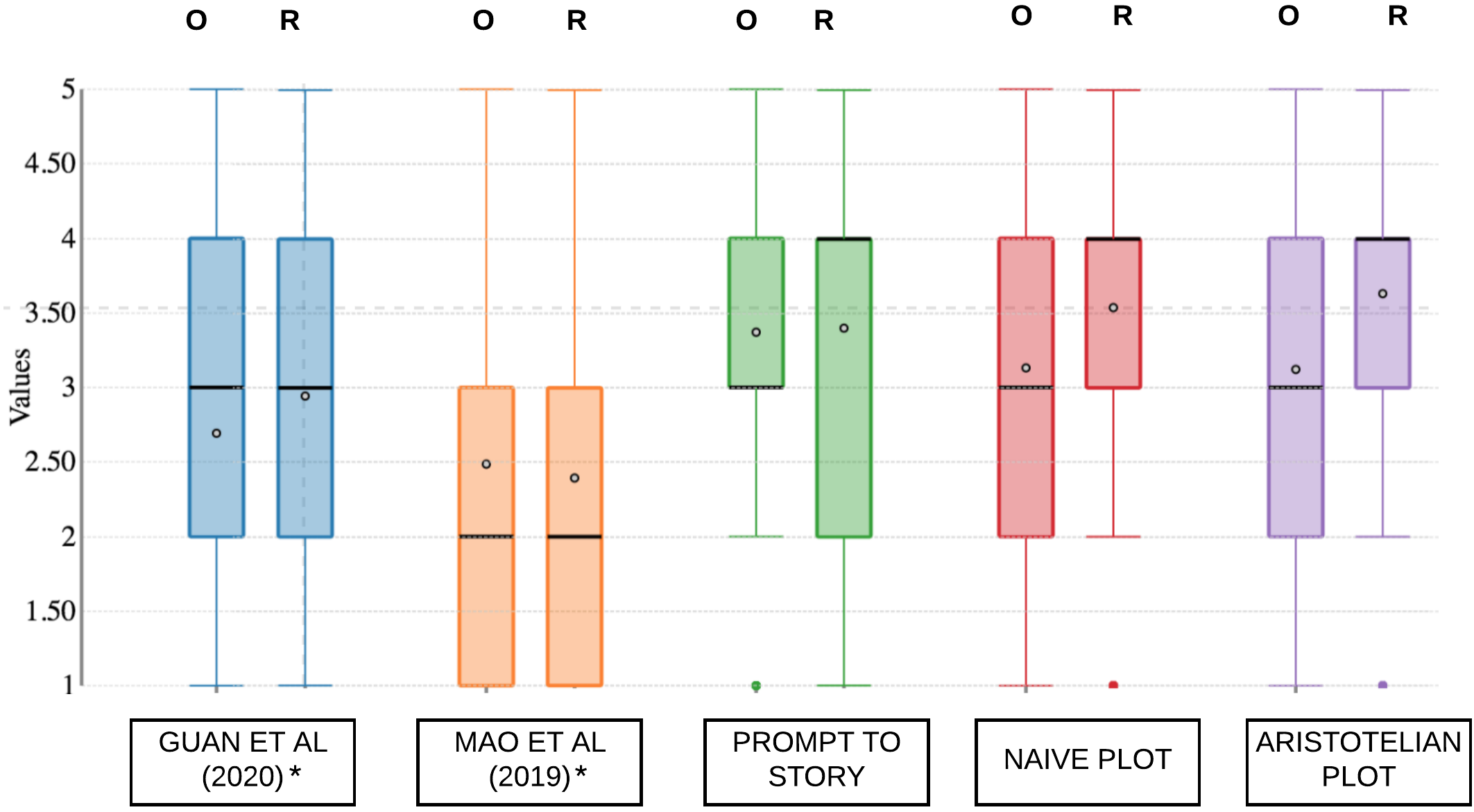}
    \caption{Likert ratings for all 5 systems scored together. O is Overall, R is Relevance, solid bars are median, dots are mean. * denotes systems with significant differences (${\alpha<0.05}$, Wilcoxon signed-rank test). The three BART based systems at right are not statistically significantly distinguishable via this setup due to central tendency bias.}
    \label{fig:likert_metrics}
    \end{subfigure} \hfill
    \begin{subfigure}[t]{0.45\textwidth}
    \includegraphics[width=\textwidth]{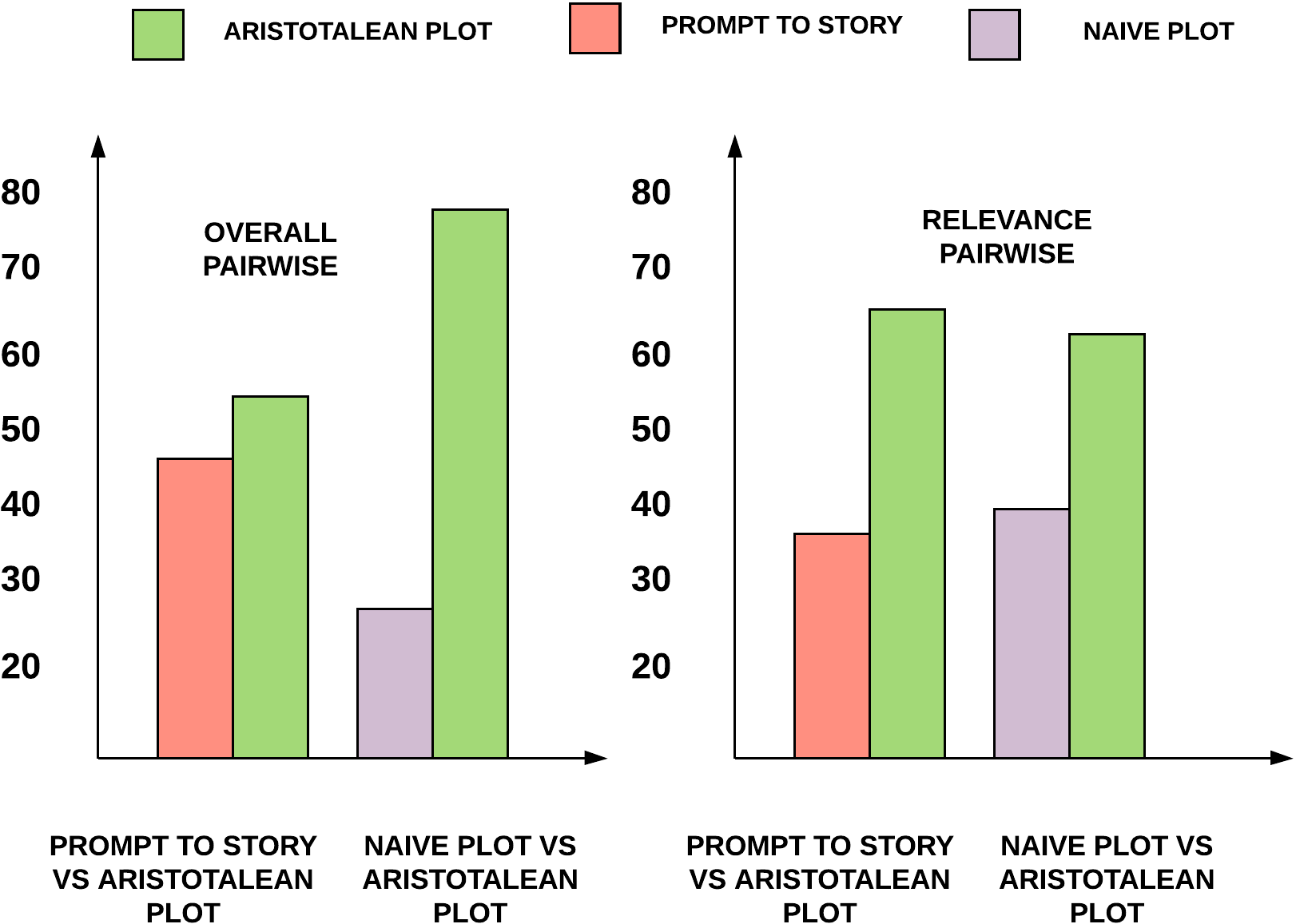}
    \caption{Pairwise win rates for the Aristotelian Plot system and two best baselines. Significant (${\alpha<0.05}$) via Wilcoxon signed-rank test.}
    \label{fig:pairwise metrics}
    \end{subfigure}
    \vspace{-.5em}
    \caption{Human evaluation for five systems on 110 prompts, with each sample evaluated by 3 judges. The scores for the likert experiment are averaged over all the instances; the pairwise comparison takes the majority vote of the three judges for each instance.}
    \vspace{-1em}
\end{figure*}

\paragraph{Plots.}
As is shown in Table \ref{tab:plot-results}, the Aristotelian plot brings the generated plot structure closer to the gold plot Vocab:Token ratio, though there is still a large gap. This improvement comes at a slight expense of number of entities per plot, which is likely because rescored plots are on average shorter. 
\paragraph{Stories.}
The combination of automatic and human evaluation for stories highlights both the coarseness of automatic evaluation metrics and the blind spots of human evaluation \cite{hashimoto2019unifying}. From just the lexical metrics, the two baseline systems of \newcite{mao2019improving} and \newcite{guan2020knowledge} would appear to have the best performance (though \newcite{guan2020knowledge} suffers in Diverse Verbs), and the three BART-based systems are indistinguishable save an improvement in Diverse Verbs with the introduction of plot structures, as observed in the plot experiments of \newcite{Fan2019StrategiesFS}.\footnote{Though we do observe a similar magnitude increase in Diverse Verbs with the introduction of plots, our baseline Prompt-to-Story model has a higher \% than their best model, reflecting the recent performance improvements in pretrained language models. We report Unique Verbs for comparability, but do not find it to be useful metric since it is not normalised by length or token count.}
However, intra-story trigram repetition reveals the weakness of the Prompt to Story system and the \newcite{guan2020knowledge} system. It also shows the extreme lack of fluency in the \newcite{mao2019improving} system, which exhibits an unnatural level of repetition characteristic of degenerate output.

\paragraph{Human Evaluation.} The human metrics in Figure \ref{fig:likert_metrics} then reinforce that the high lexical scores of the \newcite{guan2020knowledge} and \newcite{mao2019improving} systems are a result of those models sometimes deteriorating into nonsensical text, as both of those systems have extremely low human judgement scores.\footnote{Examples from each baseline system for the prompt in Table \ref{tab:first-examples} can be found in Appendix \ref{sec:appendix_story_examples}} The likert scores favor the Aristotelian Plot system with regard to relevance but favor the Prompt to Story baseline for Overall Quality - though as can be seen from the variance in \ref{fig:likert_metrics}, all three BART systems are too close together to be reliably distinguished via likert metrics. The pairwise comparisons in Figure \ref{fig:pairwise metrics} do differentiate the three systems with strong statistical significance, showing the superiority of Aristotelian Plot over both the Prompt to Story and Naive Plot systems with respect to both Relevance and the Overall Quality of the final stories. 

\section{Analysis}
We analysed the patterns in reported user confidence and found that 8\% of prompts are low user confidence (${<}$3) and 8\% are high confidence (${>}$4.5), so we look further into these as examples where the top three systems are minimally and maximally distinguishable. We include examples of outputs for these prompts for all three models in Appendix \ref{sec:appendix_conf_examples}.

For Overall Quality, there are no low confidence titles, and the Aristotelian plot system is preferred for all of the high confidence ones. 
In Relevance, the low confidence prompts have split win-rates across all models (essentially random) and the prompts show that all systems struggle with the meta-level concepts. The Writing Prompts dataset varies from  concrete (\textit{A story of a cat who saves a girl from her depression}) to prompts requiring other types of meta-knowledge. Prototypical low confidence titles are of two forms: 1) \textit{A day in this life in a world where everything is written in poetry}, and 2) \textit{Write a story where I can't tell whether the protagonist is a hero or a villain}. 
The type 1 stylistic instruction prompt requires knowledge about the distinction between instructions for style and content, and all models fail. The type 2 meta-content type of prompt requires a finer level of control of the structure of the plot and story than the Aristotelian system or any of the other models can manage.
This kind of prompt presents an interesting case for future story generation work, as it is simple enough for human authors to be popular on a forum, but far beyond the capabilities of current language models. At the high confidence Relevance prompts, the Aristotelian system wins all but one. Those stories highlight the way that adding and then improving on plot structures assists relevance by keeping a story on topic throughout an entire generation (see Appendix \ref{sec:appendix_conf_examples} Table \ref{tab:high_conf_relevance}). 

To assess the cases where the Aristotelian Plot did not improve over the BART baselines, we measure both word and verb incorporation rates, in Table \ref{tab:incorp_rates}\footnote{\newcite{yao2018plan} also use word incorporation rates, but theirs are not comparable to ours, as we both include verbs as their own separate metric (due to their importance in our structure) and make this an \textit{ordered} metric (rather than set intersection as they do), which is necessary because our generated stories are much longer}. These measure the Levenshtein distance over the sequence of plot words (excluding verbs) or plot verbs that are included within the final story. 
While the high incorporation rates show that the story model does utilize the plot, there is a gap between the current utilisation and the possible upper bound. The focus of our work is on improving plot generation, but we hypothesise that modifications to the story model to improve incorporation rates would further widen that performance gap between the three systems, as it will give the plot more influence over the story surface realisation and ensure that plot improvements appear downstream.  


\begin{table}
\centering
\small
\begin{tabular}{|@{\ \ }l@{\ \ }|@{\ \ }l@{\ \ }|@{\ \ }l@{\ \ }|}
\hline
System & Word Incorp \% & Verb Incorp \%  \\
\hline
Naive Plot & 75.5 & 75.1  \\
\hline
Aristotelian Plot & 72.4 & 72.0  \\
\hline

\end{tabular}
\caption{Metrics for incorporation rates for stories.}\smallskip
\label{tab:incorp_rates}
\vspace{-1em}
\end{table}

\section{Related Work}
\label{sec:related_work}
\paragraph{Story Generation without Plots.}
Diverse efforts have focused on generating stories. \newcite{fan2018hierarchical} re-purpose an approach for Neural Machine Translation to translate from prompt to a story via Convolutional Seq2Seq models. \newcite{guan2020knowledge,mao2019improving} use a similar approach, however they incorporate structured commonsense knowledge from external datasets or knowledge bases to improve a story generated from a prompt. \newcite{peng2018towards} add control to the story ending valence.


\paragraph{Story Generation with Plots.}
\newcite{riedl2010narrative} use refinement search
as a technique to balance between character and plot for solving the narrative generation problem. \newcite{li2013story} use plot graphs for story generation that model the intended logical flow of events in the virtual world as a set of precedence constraints between plot events. \newcite{martin2018event} decompose the problem of story generation into generation of successive events (event2event) followed by generation of natural language sentences from events (event2sentence). \newcite{Ammanabrolu2020StoryRE} follow up this work by comparing five different models for event2sentence realisation, and find all to have different weaknesses. \newcite{yao2018plan} improve their LSTM-generated ROCStories by extracting keyword-based plot-like structures, or \textit{storylines}, and using these in a pipeline to generate storylines and then stories. \newcite{Fan2019StrategiesFS} experiment with numerous techniques for representing story plots on the WritingPrompts dataset, and find Semantic Role Labelling with Entity Anonymization and Compression to work best. More recently, \newcite{Tambwekar2019ControllableNS} propose a goal-driven reinforcement learning approach to plot generation, and \newcite{chen2019learning} propose a latent variable model to learn how to generate outlines for neural story generation.  
\paragraph{Learning Story Aspects via Rescoring.}
\newcite{holtzman2018learning} generate continuation sentences from context sentences, and introduce using a mix of collaborative discriminators that each learn one Grician maxim of conversation, and use them to rerank their RNN story model output. \newcite{GoldfarbTarrant2019PlanWA} use those discriminators with the system of \newcite{yao2018plan} as part of a collaborative story writing task with an LSTM and human writers. However, none of them apply rescorers to the plot.
There is no other work that uses discriminators or rescores for plot structures or to try to train them based on different principles.  

\section{Conclusion and Future Work}
We have shown that Content Planning via an interim \textit{plot structure} representation can be combined with the use of rescoring models to inject Aristotelian story-writing principles into the plot. We found that this results in stories that are both more relevant and higher quality than stories that are generated directly from prompts or that use plots without Aristotelian rescoring. Our findings also suggest future work on additional ways to incorporate story principles into plot generation. Although our Aristotelian plots improved over the naive plot, there remains gaps in quality between generated and gold plot structures. There is also further work to be done in investigating what models are best able to incorporate plots, which would enable plot improvements to be even more effective.

\section*{Acknowledgments}
We thank the anonymous reviewers for their feedback, Adam Lopez for his excellent editing suggestions, and Jinrui Yang for her early SRL prototyping.
This work is supported by Contract W911NF-15-1-0543 with the US Defense Advanced Research Projects
Agency (DARPA). 
\bibliographystyle{acl_natbib}
\bibliography{anthology,emnlp2020}

\clearpage
\appendix
\section{Appendices}
\label{sec:appendix}

\subsection{Data Quality in Mechanical Turk Studies}
\label{sec:appendix_data_qual}
We require that Turkers doing pairwise story comparison report their confidence in their decisions and are clear that this makes no difference in their remuneration, and then we use patterns of confidence scores to find areas where models are very distinguishable and areas where they are not.

We additionally create manual True-False questions for each prompt used in human evaluation to be used as an attention check. We experimentally found that extractive attention checks (\textit{this keyword is in the title}) were ineffective, but manually created simple ones such as \textit{The protagonist is a child} were very effective at filtering out poor quality responses. 15\% of all responses failed attention checks, and we excluded that data and reran those human evaluations until we had a full dataset. We did additionally verify that none of our questions were overly difficult by manually reviewing all attention checks that ${<80\%}$ of respondents passed.

\subsection{Dataset Statistics}
\label{subsec:dataset_stats}
The Writing Prompts dataset contains 303358 pairs of prompts and stories. Stories are already tokenized in the available dataset \cite{fan2018hierarchical}. Like \cite{fan2018hierarchical} do we truncate to a maximum length of 1000 words per story (with an average of ~700 words) and replace words that appear less than 10 times with an UNK token.

We split the dataset into Train/Validation/Test splits of 65/10/10 for fine-tuning Language Models, with the remaining 15\% split into 10 for fine-tuning Rescoring models, and 5 for training Mixture Weights and generally running metrics and tests. 
\subsection{Implementation Details}
\label{subsec:appendix_impdet}
We fine-tune BART with same hyperparameters mentioned in \footnote{\url{https://github.com/pytorch/fairseq/blob/master/examples/bart/README.summarization.md}}  with the exception of MAX-TOKENS (size of each mini-batch, in terms of the number of tokens.) being 1024 for us. For decoding we generate plots and stories from our models using a top-k random sampling scheme \cite{fan2018hierarchical}. At each timestep, the model generates the probability of each word in the vocabulary being the likely next word. We randomly sample from the k = 5 most likely candidates from this distribution. We also use a softmax temperature of 0.7.

\subsubsection{Fine-Tuning BART for Prompt to Plot}

\begin{enumerate}
    \item{\textbf{No of Parameters:}} For BART we use the BART large checkpoint (400M parameters) and use the implementation by FAIRSEQ \cite{ott2019fairseq}
     \footnote{\url{https://github.com/pytorch/fairseq/tree/master/examples/bart}}
    \item{\textbf{No of Epochs:}} We fine-tune pre-trained BART for 19 epochs for FTB model  keeping all default hyper-parameters as mentioned in \url{https://github.com/pytorch/fairseq/blob/master/examples/bart/README.summarization.md} with the exception of $MAX\_TOKENS=1024$ and $UPDATE\_FREQ=16$.
    \item{\textbf{Training Time:}} Our training time is 32 hours
    \item{\textbf{Hardware Configuration:}} We use 2 RTX 2080 GPU (11GB)
\end{enumerate}

\subsubsection{Fine-Tuning BART for Plot to Story}

\begin{enumerate}
    \item{\textbf{No of Parameters:}} For BART we use the BART large checkpoint (400M parameters) and use the implementation by FAIRSEQ \cite{ott2019fairseq}
     \footnote{\url{https://github.com/pytorch/fairseq/tree/master/examples/bart}}
    \item{\textbf{No of Epochs:}} We fine-tune pre-trained BART for 17 epochs for FTB model keeping all default hyper-parameters as mentioned in \url{https://github.com/pytorch/fairseq/blob/master/examples/bart/README.summarization.md} with the exception of $MAX\_TOKENS=1024$ and $UPDATE\_FREQ=16$.
    \item{\textbf{Training Time:}} Our training time is 6 days due to limited GPU RAM and since stories and plots are large sequence of tokens.
    \item{\textbf{Hardware Configuration:}} We use 1 RTX 2080 GPU (11GB)
\end{enumerate}

\subsubsection{Fine-Tuning roBERTa-large for Discrimimators}
\begin{enumerate}
 \item{\textbf{No of Parameters:}} For BART we use the roBERTa large checkpoint (355 parameters) and use the implementation by FAIRSEQ \cite{ott2019fairseq}
     \footnote{\url{https://github.com/pytorch/fairseq/blob/master/examples/roberta/}}
    \item{\textbf{No of Epochs:}} We fine-tune pre-trained roBERTa 7 epochs for relevance discriminator model, 5 epochs for entity discriminator models, 4 epochs for inter-event, 2 epochs for intra-event and 10 epochs for intraV-event discriminator keeping all default hyper-parameters as mentioned in \url{https://github.com/pytorch/fairseq/blob/master/examples/roberta/README.glue.md} with the exception of max-tokens$=1024$ and lr-scheduler=fixed.
    \item{\textbf{Training Time:}} Our training time is 10 hours for relevance discriminator , 4 hours for entity discriminator ,4 hours for inter-event, for 2 hours for intra-event 12 hours for  intraV-event discriminator
    \item{\textbf{Hardware Configuration:}} We use 4 RTX 2080 GPU (11GB)
\end{enumerate}

\subsection{Hyper-Parameters for Mixture Weight Tuning}
\label{sec:appendix_mix_weight}
Mixture weights are tuned with a held out validation set of 10,000 samples. The models train for 3 epochs, but all converge in ~1 epoch, which takes 24 hours on 1 RTX 2080 GPU (11GB). We use SGD at each step, with learning rates set to 0.001. 

\subsection{SRL-Event Representations}
\label{sec:appendix_stopverbs}
We compress event-representations by removing all verb arguments beyond args 0,1,2. 

We exclude the following verbs when extracting plot events:{\tt"is", "was", "were", "are", "be", "\'s", "\'re", "\'ll", "can", "could", "must", "may", "have to", "has to", "had to", "will", "would", "has", "have", "had", "do", "does", "did"}

\section{Example Training Data for Rescorers}
\label{app:disc_training_examples}
Table \ref{tab:disc_examples} shows example positive and negative training data for all rescorers for the prompt \textit{People gather around a camp fire to tell the story of a legend , who actually sits among them}. Training data is excerpted for display, but otherwise unedited.

 \begin{table*}[t]
 \caption{Example (excerpted) training data for all rescorers for the prompt \textit{People gather around a campfire to tell the story of a legend, who actually sits among them}. The Relevance rescorer includes one sentence of plot in the given context, the Entity rescorer's context is the entire plot up to a randomly chosen entity.}
 \centering
 \small
 \begin{tabular}{l|l|p{4cm}|p{4.5cm}|p{4.5cm}}
 \multicolumn{2}{l|}{} & Context & Positive & Negative\\ 
 \hline
  \multirow{3}{*}{Event} & Inter & 
  \multirow{3}{4cm}{People gather around a camp fire to tell the story of a legend , who actually sits among them .} & {\tt <<A1> ent 0 orange glow <V> stood <A2> ent 6 night  </s>
<A1> ent 3 <V> emanating <A2> ent 3  </s> <A0> ent 2 <V> felt <A1> the cold <A2> ent 2 their backs \# <A0> ent 2 <V> faced <A1> ent 3  </s><A1> ent 2 eyes <V> stayed <A2> upon the saving light \# <A0> ent 4 <V> stared  </s>...}  & {\tt <<A1> ent 3 <V> emanating <A2> ent 3  </s>  
 <A1> ent 8 <V> grew <A2> quieter , darker  </s>  
 <A2> ent 5 some <A1> ent 5 <V> came \# <A0> a bearded , old man <V> drawing <A1> ent 11 <A2> close \# <A1> ent 13 <V> burn  </s>
 <A0> orange <V> glow \# <A1> ent 1 <V> sat \# <A1> ent 1 <V> paralyzed  </s> ...} \\ \cline{4-5}
 & Intra & & {\tt <A0> ent 2 <V> felt <A1> the cold <A2> ent 2 their backs \# <A0> ent 2 <V> faced <A1> ent 3  </s>
<A1> ent 2 eyes <V> stayed <A2> upon the saving light \# <A0> ent 4 <V> stared  </s>...} & {\tt <A0> ent 2 <V> faced <A1> ent 3  \#  <A0> ent 2 <V> felt <A1> the cold <A2> ent 2 their backs </s>
<A0> ent 4 <V> stared  \#  <A1> ent 2 eyes <V> stayed <A2> upon the saving light </s>...} \\ \cline{4-5}
 & Verb & & {\tt <A0> ent 9 <V> roamed <A1> the woods \# <A0> ent 9 <V> consumed <A1> ent 6 of the night  </s>  <A0> The wind <V> began <A1> to blow with cold intention \# <A1> The wind <V> blow \# <A0> ent 7 <V> danced \# <A1> ent 7 <V> shimmered \# <A1> moonlight <V> began} & {\tt <A0> ent 9 <V> consumed <A1> the woods \# <A0> ent 9 <V> roamed <A1> ent 6 of the night  </s>  <A0> The wind <V> shimmered <A1> to blow with cold intention \# <A1> The wind <V> began \# <A0> ent 7 <V> danced \# <A1> ent 7 <V> blow \# <A1> moonlight <V> began} \\ 
 \hline
 \multicolumn{2}{l|}{Entity} & {\tt <A0> ent 0 <V> saw <A1> the light of a campfire </s> <A1> ent 2 <V> laying <A2> there </s> <A1> horses <V> surrounding <A2> ent 2 \# <A1> light <V> bouncing </s> <A0> ent} & \textbf{0} & \textbf{3} \\
\hline
 \multicolumn{2}{l|}{Relevance}  & People gather around a camp fire to tell the story of a legend , who actually sits among them .{\tt <EOT> <V> masked <A0> ent 0 \# <A0> ent 0 <V> rode </s>} & {\tt <A0> ent 0 <V> saw <A1> the light of a campfire  </s>  <A1> ent 2 <V> laying <A2> there  </s>  <A1> horses <V> surrounding <A2> ent 2 \# <A1> light <V> bouncing  </s>  <A0>...}
 & {\tt <A0> ent 2 <V> asks <A2> ent 0  </s>  <A1> I <V> ' <A2> sorry \# <A0> I <V> think <A1> ent 0 can help you \# <A0> I <V> help <A1> ent 0  </s>   </s>  <V> colored <A1> toys  </s>...} \\
\hline
 \end{tabular}
 \label{tab:disc_examples}
 \end{table*}

\section{Example Outputs For High and Low Confidence Scores}
\label{sec:appendix_conf_examples}
All examples in Tables \ref{tab:high_conf_overall}, \ref{tab:high_conf_relevance}, and \ref{tab:low_conf_relevance} are unedited save removal of special separation tokens and detokenization. Line breaks are added by the model.
\begin{table}
\centering
\small
\caption{High Confidence: Overall Quality (score 4.67 out of 5)}\smallskip
\begin{tabular}{p{7.2cm}}
\textit{Prompt:} Earth has gained a new season, unlike anything currently known \\
\hline
\textit{Aristotelian Plot System Generated Story:} \\
The Earth is more beautiful than it thought! It's the most beautiful planet. The sun was shining bright, and the air was crisp, yet the air itself was cold. Everything was so bright. It was the perfect day for earth to be beautiful. \\ Everyone on the planet Earth had lived a long and happy life, this planet has been home to the human race for thousands of years. The humans were a species that had been able to survive through Earth's harsh environment. the humans could survive in these harsh conditions. But one day Earth had a new life, and we were the first to notice it. In the middle of the night, we could see something, a small, dark cloud. A small, white cloud, it's size was almost invisible. Then that cloud grew. And it grew...\\
The sun, shining bright.. There was no sun. All the light was black and it covered the sky. Every day the sky was a dark blue. No matter the time of day, it never seemed like the perfect time, it was a beautiful day. For a new season to begin. Yet no one could know why. Why did we not have a new cycle?" \\
\hline
\textit{Naive Plot System Generated Story:} \\
``It's not a new day, John. '' \\
`` What? What is it? You ca n't see?''\\
The man was silent.\\
He said, ``It'd be the first time we 've all seen its as we know it, but it's the first we're seen in over a million years. It is a new world, one with no sun. ''\\
A few seconds of silence.`` Its not a New Day.'' the man said.\\
''It're a new season,'' said the woman. We' ve seen them all. I mean, we all know by now. You know, I 've- seen, you' ve always seen- ''\\
``We'll see us. But you'll be the First, you were the first one to see, we'll all be the Last, we will all be, we have all been the first of our species. There is no one more, no one is better than me.'' ''The woman was right. \\
\hline
\textit{Prompt to Story Generated Story:} \\
I was in my bed, in my own room, in the middle of a forest, when the world stopped.\\ 
It was the first time in my life that I had felt I had a sense of peace. My family and me, myself, were gone, we had all been taken from us. The only thing our family had left were the three children and my world, but my wife and I were the only survivors. There were we, the three of us. But ourselves, our children were all gone. And the only reason why me and our family survived was because ours had become ourselves. It was as if both of us were one with the world. As all the children were gone and the adults had left myself to fend for myself, you and your family. No longer were you and I to be each other. Now we were just two. All our memories of us, all ours. Nothingness. Only all of us, and my children, each of us... \\
\hline
\\
\\
\\
\\
\\
\\
\\
\end{tabular}
\label{tab:high_conf_overall}
\end{table}

\begin{table}
\centering
\small
\caption{High Confidence: Relevance (4.86 out of 5)}\smallskip
\begin{tabular}{p{7.2cm}}
\textit{Prompt:}  Evolution has gone a little differently. Humans have poison and some way to transmit it.\\
\hline
\textit{Aristotelian Plot System Generated Story:} \\
So the poison is in your genome, ''said the scientist.\\
`` I have n't even it's looked at you yet Mr. Doctor!'' \\ 
``Yes sir! ''\\
`` And its in my DNA. '' said the man. \\
``Well, you must be the first human, to have developed this. I mean, you are.'' He said,`` It's not a question of if but when You. It'd be a question if your DNA would even be the same. ''\\
The man looked him in the eye. ``Well Mr. Doctor...''\\
`` Yeah you are the first, I mean sir, ''he said. * I am not sure he was. But he is who i am. Do you see that? You and I are not the first species to have it, but this one has come from the beginning of this.''. The human genome of the virus had the most advanced species known to have had It. And it was spread by a species that had the best genetic makeup in the world. A species which was a virus spread by the human race. Now, you and I were not the best of the best at it, because we did n't have the ability to develop a species who could spread the toxin. We were the best. This is why we live in this place now. Because we can live in a world with no poisons. You know we 've killed off each other, but now we live our lives, you all know. Just like myself, Mr. President, I do n't know how long it took. They live by. Well, it was about a thousand years ago. In the same world we come from, the human genome spread by The human race, and spread by that same process.  \\
\hline
\textit{Naive Plot System Generated Story:} \\
`What? People say that they know what they're saying. I know, but the ones who kill? I know. The ones who do the killing! ''\\
`` Yeah I know! They're all dead! All of them, the ones you know about! That's what you say... but what? That we're gon na kill them? You see, the one who's dead, are the ones that you 've killed. Not me. It's us.'' They are my brothers and sisters, after all. \\
The one's that will die, the people who we kill, the guys and girls that you guys and boys kill for. \\
\hline
\textit{Prompt to Story Generated Story:} \\
We are the most dangerous species in history, yet, we do not know why. We have been hunted for thousands of years. It was the same for our ancestors, they were their prey, and the same fate for it. \\
Our species evolved to be more intelligent than the others, and so they were able to survive. But the human race was n't able to adapt. So they were forced to adapt to the world themselves, and the first few years were difficult. There were no predators to eat, no predators that could eat us, no prey that could be killed, no one to protect them. All the predators had been wiped out and their prey was no longer needed. The population was so small, so few that no predator could take on the entire world. And so they died. No one was left to fight. Just the humans. That is until one of them, an old man named the man who was the most intelligent and powerful of all of them... A few of them were his friends. Some were his wife, others his kids, and some even he was a friend. Then the humans came to their deaths, they were not as strong as them. ''\\
``And we were the last to go.'' ''The man continued." \\
\hline
\end{tabular}
\label{tab:high_conf_relevance}
\end{table}

\begin{table}[t]
\centering
\small
\caption{Low Confidence: Relevance (2.8 out of 5)  }\smallskip
\begin{tabular}{p{7.2cm}}
\textit{Prompt:} Write a story where I can't tell whether the protagonist is a hero or a villain.  \\
\hline
\textit{Aristotelian Plot System Generated Story:} \\
"The men seemed a little more than usual, a bit. The women were n't the same. No.\\
They were all different. All of them, my wife. Each one of them were given the love of a woman, and none of them knew me as I did. If it took the people of the city to realize what they had done with the world, the world would n't have been so different...\\
But this man. He was my hero. The one who was the one who made the difference, the only thing what hadkept the women from the other women. Theone who had saved the day. And the woman was the most special. It was the woman, the one that he had loved. But who would take the fall for the save a single life, for a person who saved his wife, her life. Not a single person was going to be able to save from this. There was no one to save. Nobody. Even the police. That seemed like it to us. ``No one cared. ''. In a world where there were no people who needed help, no people that could need help. A world where those who needed for others, were the ones who would be the ones to do the work. \\
\hline
\textit{Naive Plot System Generated Story:} \\
So he's the hero. He's a good guy. Or is it?'' ``We did n't know him before. But we come in here for a meeting. What's his name? ''\\
`` We'll have a talk. Do n't go anywhere, my name is n't your name.''\\
* His name is not mine, I 'm not sure its a real name. *\\
I'M not sure if I' ve ever been told I was a hero, or just a villain. \\
``No, no, you're not. I'm not. ''\\
`` What does that even mean?'' ''The man ask me. ``What is your name? Is yours name really mine?''... \\
-- -- -- \\
The mayor of this small town is on the phone with his secretary. “We 've got a new guy. ''the woman says. A young man with a black suit and a dark suit. In the suit, he was told she was a heroine. No. \\
\hline
\textit{Prompt to Story Generated Story:} \\
I was a hero.\\
I have never seen a hero before, and me being one is something you do n't expect in your first year of high school.\\
The first thing that happened to my mind was that i was a kid, and then its a little hard to explain. The first few weeks of school were a blur. A few months went by and then it was just myself and my best friend. Then the bullying started. And then she started to do her own shit. So then it was mine, and soon it was all she did. But he did n't know what to do either. It was n't until his first few grades that it was nt until he started to talk. When we were in middle school that was when the bully started to take things out on him. That was when it all started...\\
It took us about a year before our teacher finally got the courage to tell the class to call in sick to school.. To go to the hospital. \\
\hline
\end{tabular}
\label{tab:low_conf_relevance}
\end{table}


\section{Output from Baseline systems for the first example prompt}
\label{sec:appendix_story_examples}
Output from the remaining models for the example prompt in Table \ref{tab:first-examples} of the paper: \textit{Write the most beautiful end to the universe that you can imagine}. All examples are unedited save removal of special separation tokens and detokenization.

\begin{table}[t]
\centering
\small
\caption{Output from Baseline systems for the first example prompt}\smallskip
\begin{tabular}{p{7.2cm}}
\textit{Prompt:} Write the most beautiful end to the universe that you can imagine  \\
\hline
\textit{Prompt to Story Generated Story:} \\
We were n't sure how long it would last. It was only a matter of time before something would happen and the universe would be gone. \\ 
It had been a long time since the end had come. A long time ago they had been trying to make it happen for us but we were nt sure how its going to be , or how they would do that. But now its over. Its time. The end was here and I am the last one standing. And now you will know your end. This is my last post\\ 
\hline
\textit{\newcite{guan2019story} Generated Story:} \\
The wind blotted it's path through the clouds, the wind whipping through the trees, the sky. It was a beautiful day. - The wind cut through his hair and made him stop, a tear rolling down he'd always dreamed about. ''That day. ''. ity. a dark shadow stretched out and reached down into the man's pocket. h He took a deep breath, and looked at the wind. a small flash. i'' Hey! '. 'hanging up her coat she said, walking up the street. d her legs made a sound as they crossed a busy sidewalk. b At this point, the wind took out its last breath. u'! hashing the door! 'y. y it made a loud noise and shouted at Katerina. y The noise stopped when it started to move again. z ``Oi.'... 'I'm sorry, I was just walking up. ''She stood up, and took out a small pack of cigarettes. m It kept you company. o Katerinas started to walk towards Keiran. t, Mina's voice was coming down. th! y! t Th As Tchaikovsky started to play the piano, the wind stopped, its face fell to the ground. p * * ''What in the hell are you doing.' *. The wind stopped again.. * Keirans eyes followed. 1.. y. Mina was sitting on the bench, Keirany had a smirk on her face.\\
\hline
\textit{\newcite{mao2019improving} Generated Story:} \\
The Mysterious Crawl -- -- -- -- -- -- -- -- -- -- -- -- -- -- -- -- -- -- -- -- -- -- -- -- -- -- -- -- -- -- -- -- -- - Part 1: Part 2: Part 3: Part 4: It was an exercise in Mars shoubu!!!; where one record holder just spins out, another record holder flies in and out of a hole in a water is a caterpillar. It's like the practice. Pre-secession: On one side, watches ``forever ''end in its own universe, one question is raised: The term`` end of the universe,'' or the likely future, or the child of the universe, may be brought up and translated as ``no matter what is said, it is true. ''\\
\hline
\\
\\
\\
\\
\\
\\
\\
\\
\\
\\
\\
\\
\\
\end{tabular}
\label{tab:more_first_examples}
\end{table}

\section{Mechanical Turk Study Materials \& Compensation Details}
\label{sec:appendix_mturk}
Human Evaluation consisted of 110 titles evaluated by 3 judges each, which results in 330 x 5 long stories for likert comparisons and 1320 x 2 long stories for pairwise comparisons (as there are 4 sets of pairwise comparisons). Turkers received \$12/hr, so the entire budget was \$2k.\\
Included at the end of the appendix in Figure \ref{fig:mturk_survey} are the pairwise surveys for Relevance and Overall Quality, for reference. HTML for these surveys and for the likert surveys are included in the code repository.
\clearpage

\begin{figure*}
    \caption{Example Mechanical Turk Pairwise Surveys for \textbf{Relevance} followed by Surveys for \textbf{Overall Quality}, including macro for attention check.}
    \label{fig:mturk_survey}
    \ContinuedFloat*
	\includegraphics[page=1,width=\textwidth]{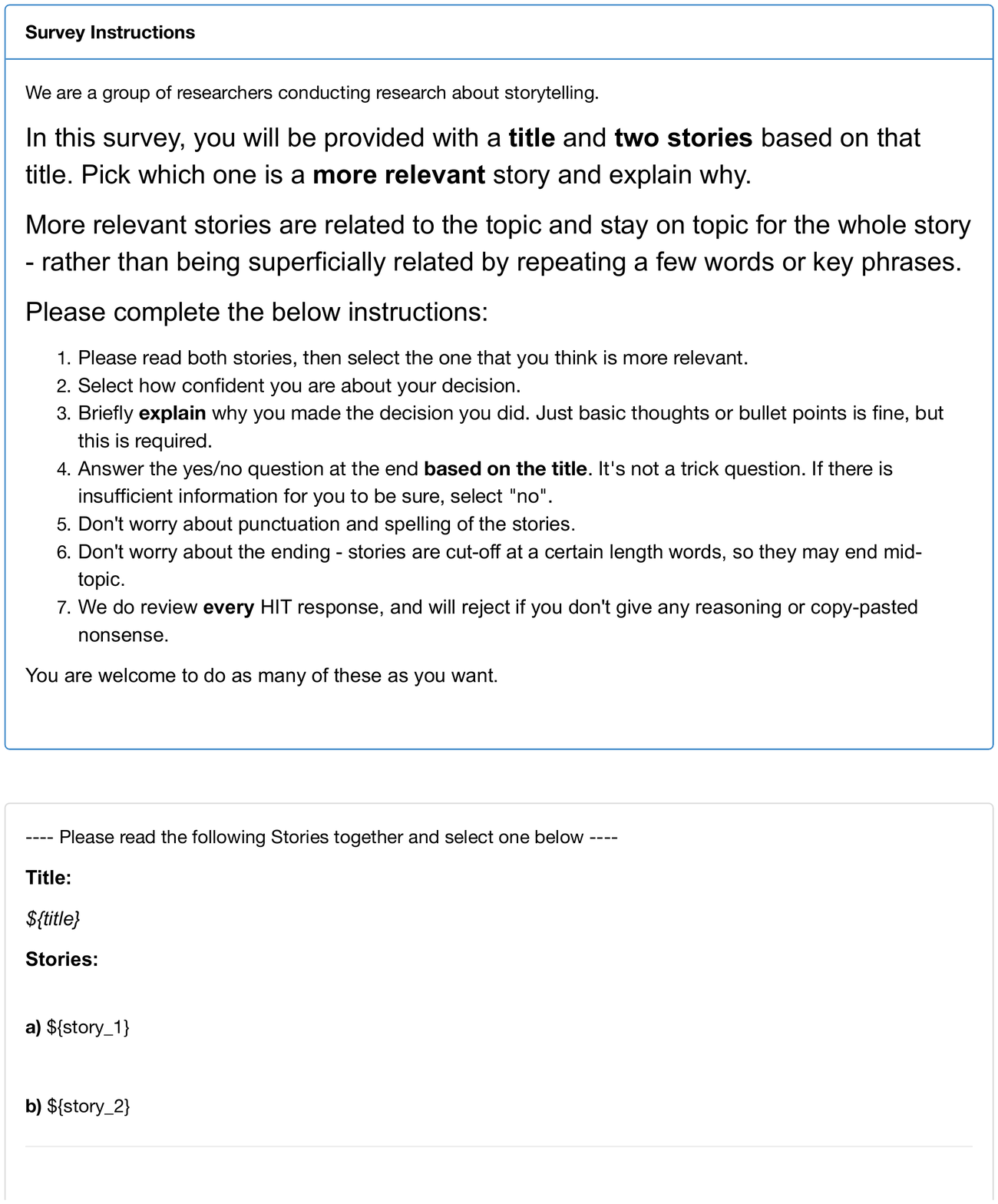}
\end{figure*}
\begin{figure*}
    \ContinuedFloat*
	\includegraphics[page=2,width=\textwidth]{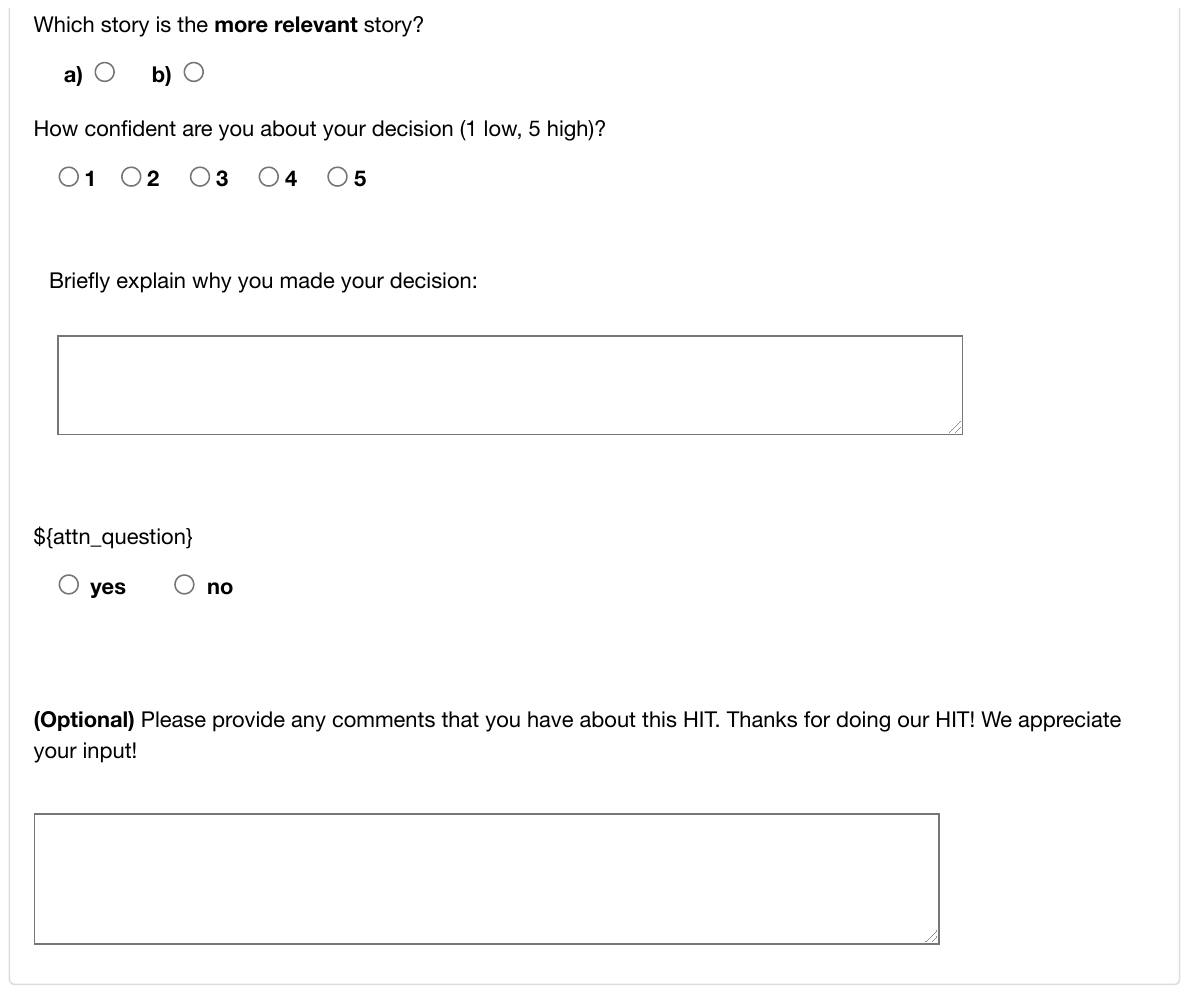}
\end{figure*}

\begin{figure*}
    \ContinuedFloat*
	\includegraphics[page=1,width=\textwidth]{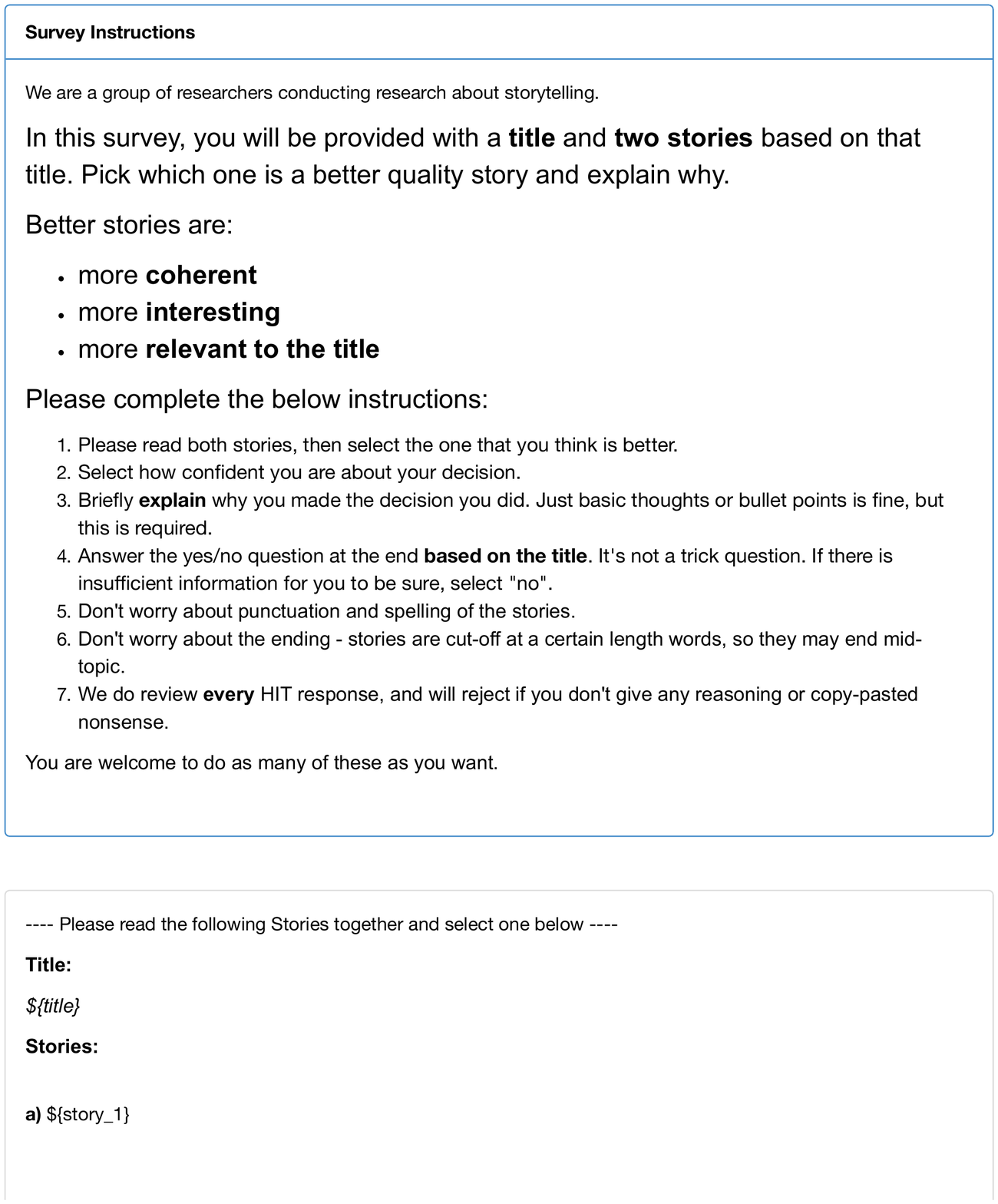}
\end{figure*}
\begin{figure*}
    \ContinuedFloat*
	\includegraphics[page=2,width=\textwidth]{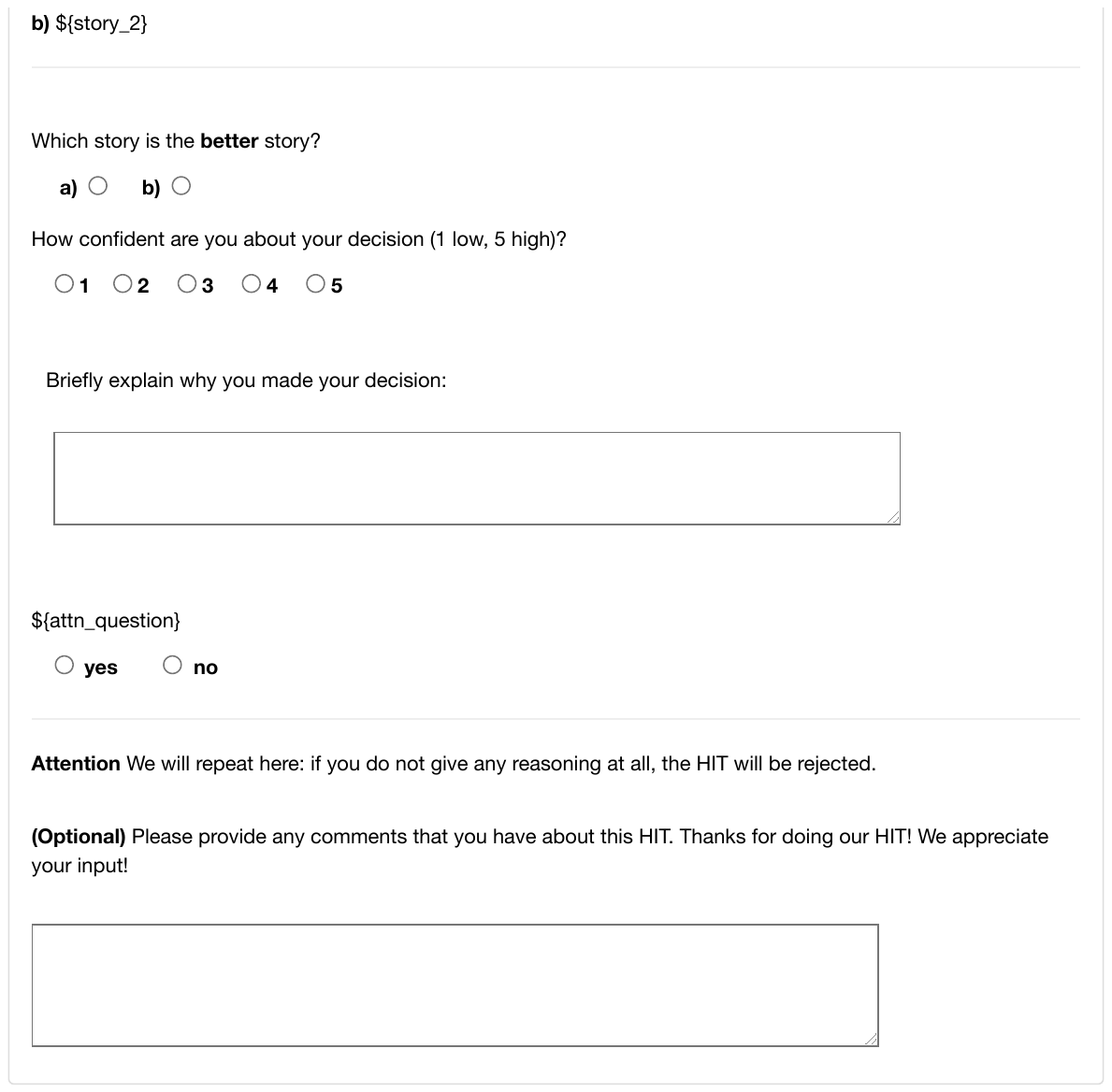}
\end{figure*}

\end{document}